\pdfoutput=1

\documentclass[11pt]{article}

\usepackage{acl}

\usepackage{times}
\usepackage{latexsym}
\usepackage{url}
\usepackage[T1]{fontenc}
\usepackage{tipa}

\usepackage[utf8]{inputenc}

\usepackage{microtype}

\usepackage{inconsolata}
\usepackage{adjustbox}
\usepackage{kbordermatrix}

\title{A Computational Model for the Assessment of Mutual Intelligibility Among Closely Related Languages}

 \author{Jessica Nieder \\
       MCL Chair \\
       University of Passau \\ 
       Passau, Germany \\ \texttt{jessica.nieder@uni-passau.de}
       \And 
    Johann-Mattis List \\
    MCL Chair / DLCE \\
    University of Passau / MPI-EVA \\
    Passau / Leipzig, Germany \\
    \texttt{mattis.list@uni-passau.de}}

\begin{document}
\maketitle
\begin{abstract}
Closely related languages show linguistic similarities that allow speakers of one language to understand speakers of another language without having actively learned it. Mutual intelligibility varies in degree and is typically tested in psycholinguistic experiments. 
To study mutual intelligibility computationally, we propose a computer-assisted method using the Linear Discriminative Learner, a computational model developed to approximate the cognitive processes by which humans learn languages, which we
expand with multilingual semantic vectors and multilingual sound classes. We test the model on cognate data from German, Dutch, and English, three closely related Germanic languages.  We find that our model's comprehension accuracy depends on 1) the automatic trimming of inflections and 2) the language pair for which comprehension is tested. Our multilingual modelling approach does not only offer new methodological findings for automatic testing of mutual intelligibility across languages but also extends the use of Linear Discriminative Learning to multilingual settings. 
\end{abstract}

\section{Introduction}
Speakers of a given language can often partially comprehend other languages in the same language family. This mutual intelligibility has been demonstrated to be dependent on several linguistic variables, such as phonological, orthographic or lexical similarity, and extralinguistic factors, such as the amount of previous exposure to or the attitude towards the other language \citep{gooskens_swarte_2017, Gooskensetal2015, heeringa_swarte_schüppert_gooskens_2014}. In addition, the phonetic similarity between words expressing similar meanings  has been shown to be a major factor driving cross-linguistic mutual intelligibility \citep{gooskens2018mutual}. Phonetically and semantically similar words are often called \emph{cognates} in studies on mutual intelligibility, foreign language learning, and bilingualism \citep{Squires2020}. Originally, however, the term denotes words inherited from the same ancestral language in genetically related languages \citep{List2016f}. Although cognates in the original sense often exhibit phonetic and semantic similarity across related languages, they do not necessarily do so, and words can also be similar in pronunciation and meaning due to other factors, including -- most importantly -- intensive borrowing, and -- to a much lower degree -- different kinds of sound symbolism (see \citealt[87]{Casad1987} for more details on the difference between mutual intelligibility and genetic relationship).

Due to a focus on the abilities of language users, research on mutual intelligibility often involves experimental studies with different groups and numbers of 
participants. Experiments are diverse, usually consisting of certain comprehension tasks. Experimental studies show some general limitations, in so far as uniform methods are rarely used~ (1), finding participants with a minimum or no exposure to the test language is difficult~(2), and comparing several languages simultaneously is a time- and resource-consuming effort~(3) \citep{gooskens_swarte_2017, TANG2009709}. 
\citet{gooskens_swarte_2017} present a large-scale study on mutual intelligibility of five Germanic languages using a \emph{Cloze Test}, i.e. a written or audibly presented text in the target language with gaps that need to be filled in.
However, they report a substantial loss in the number of participants when testing \emph{inherent intelligibility}, the ability to comprehend the target language with no or little previous exposure \citep{gooskens_swarte_2017}. In an ideal setting with zero exposure to the target language, inherent intelligibility captures how comprehensible the target language is based on structural similarities only. This, in turn, would offer insights into what linguistic structures give rise to mutual intelligibility without extralinguistic or other language exposure-based interference. In reality, the goal of finding participants with no or a minimum of exposure to certain languages is an almost impossible requirement to fulfill due to the status of some languages of being a common \emph{lingua franca} \citep{gooskens_swarte_2017, Hongyan2017}.

\begin{figure*}[tb]
    \centering
    \includegraphics[width=0.9\textwidth]{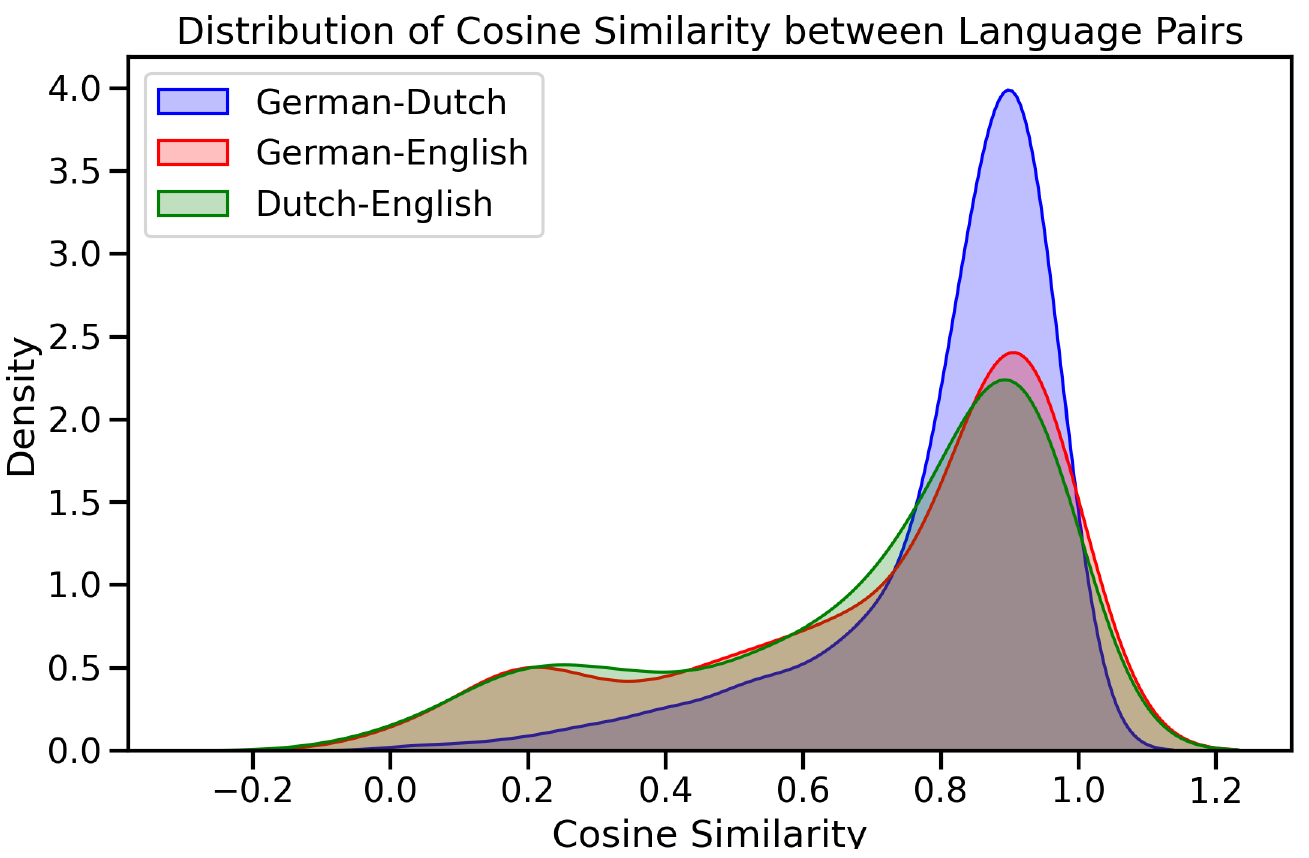}
    \caption{Distribution of cosine similarity scores between language pairs for all cognate triplets. Note that smoothing of the distribution results in values exceeding 1.0.}
    \label{fig:cosinesim}
\end{figure*}

In this study we propose a computer-assisted method couched in the discriminative lexicon framework by \citet{baayen_LDL2019} to assess mutual intelligibility in Germanic languages. By focusing on computational methods instead of human subjects we can overcome the mentioned limitations. Our proposed model does not involve the recruitment of participants, there are no extralinguistic factors nor target language exposure involved in training. 
We offer a uniform method that can be adapted to various language families and lead to new insights into intelligibility based on a careful selection of linguistic factors that are involved in language comprehension.

\section{Linear Discriminative Learning}
With the discriminative lexicon framework (DL), \citet{baayen_LDL2019} propose a model of language processing that explores the cognitive mapping mechanisms involved in language learning. Language comprehension is understood as a mapping of phonological forms onto meaning \citep{baayen_LDL2019}. 
Mathematically, it is implemented as multivariate multiple regression in the Linear Discriminative Learner (LDL) model. Given a phonological matrix \emph{$C$} and a semantic matrix \emph{$S$}, the comprehension matrix \emph{$F$} is obtained by post-multiplying \emph{$C$} with \emph{$F$}: \emph{$CF = S$}. The \emph{$F$} matrix then specifies the associaton weights between all phonological cues and all semantic dimensions \citep{chuang2023vector}. Multiplying \emph{$C$} with \emph{$F$} finally predicts the semantic vector  \emph{$\hat{S}$} for all input word forms that can be used for evaluating comprehension accuracy of the model. Computationally, the LDL model conceptualizes language comprehension as a simple artificial neural network directly connecting phonological and semantic vectors without any hidden layers \citep{niederetal_2023, chuang2023vector}. 
In this study, we make use of LDL to explore the mutual comprehension of the Germanic languages Dutch, German and English based on a cross-language learning setting \citep[see also][for another multilingual approach using LDL]{chuangetal2021multilingual}. As phonological input we use cognate sets from all languages. 
For the semantic matrix, we opted for the multilingual ConceptNet Numberbatch word embeddings \emph{version 19.08} from \citet{speer2017conceptnet} that offer the possibility to directly compare the meaning of cognate concepts. 

\section{Materials and Methods}
\subsection{Dataset of German Cognates}
We use cognate sets derived from Kluge's etymological dictionary in a rather recent, updated edition \citep{Kluge2002}. From the etymological dictionary of German, we hand-selected 340 entries that had reflexes in Dutch, German, and English with their proto-forms in Proto-Germanic, added phonetic transcriptions, and provided phonetic alignments by annotating the data with the help of the \emph{EDICTOR} tool \citep{EDICTOR}. 

In order to ease data sharing and reuse, the etymological dataset was shared in the formats recommended by the Cross-Linguistic Data Formats initiative \citep{Forkel2018a} using the workflow developed for the construction of the Lexibank repository (\url{https://lexibank.clld.org}, \citealt{List2022e}). This means in this specific case that languages are linked to Glottolog (\url{https://glottolog.org}, \citealt{Glottolog}) and that the individual speech sounds employed in the phonetic transcription we provide follows the Cross-Linguistic Transcription Systems (CLTS, \url{https://clts.clld.org}, \citealt{CLTS}). CLTS is a reference catalog for speech sounds which provides a standard transcription system that defines a subset of the International Phonetic Alphabet \citep{IPA1999} as a standard \citep{Anderson2018}, which has by now been mapped to several datasets providing phoneme inventory data \citep{Anderson2023a} and also underlies most data in Lexibank.

\subsection{Multilingual Semantic Vectors}
For semantic vectors, we used the multilingual ConceptNet Numberbatch word embeddings \emph{version 19.08} from \citet{speer2017conceptnet}. 
The ConceptNet Numberbatch word embeddings did not provide any data for the Dutch word form \emph{beukeboom} `beech', thus we deleted the German and English counterparts from the data resulting in a set of 339 cognates in total.
To ensure that the embeddings capture semantic similarites of the cognate dataset, we computed the cosine similarity for each word triplet across the languages. Figure \ref{fig:cosinesim} shows the distribution of cosine similarity values between language pairs. While the peaks for all language pairs are located at around 0.9, indicating an overall high semantic similarity of the word embeddings for cognate triplets, 
some of the German-English data and Dutch-English data
is distributed over a lower cosine similarity range (green and red curve). This results in less concentrated peaks for these language pairs. From this we can conclude that German vs. Dutch cognates are semantically more similar than German vs. English or Dutch vs. English cognates. 

\begin{table}[b!]
\centering
\resizebox{\columnwidth}{!}{%
\begin{tabular}{lll}
\hline
& \textbf{English} & \textbf{German} \\\hline
\textbf{Word} & drink & trinken \\
\textbf{IPA} & \textipa{d R I N k} & \textipa{t R I N k @ n} \\
\textbf{IPA (trimmed)} & \textipa{d R I N k} & \textipa{t R I N K @} \\
\textbf{Sound Classes} & \texttt{T R V N K} & \texttt{T R V N K V N}\\
\textbf{Sound Cl. (trimmed)} & \texttt{T R V N K} & \texttt{T R V N K V} \\\hline
\end{tabular}}
\caption{Exemplary data representation for English and German with full forms vs. trimmed forms and sound class representations.}
\label{tab:datarep}
\end{table}

\subsection{Multilingual Sound Classes}
Scholars have proposed to test mutual intelligibility by representing word forms in phonetic transcriptions and measuring string similarity for words that express the same meaning \citep{Tang2007}. This approach to 
intelligibility has, however, the disadvantage of not being able to test for \emph{asymmetric forms} of intelligibility by which speakers of one language can understand speakers of another language more properly than vice versa. 
For our model-based approach, we need a more abstract -- phonetically broader -- representation of speech sounds
that allows us to capture broad phonetic similarities in a multilingual setting. Taking inspiration from computational approaches in historical linguistics, we decided to represent word forms with \emph{sound class models}. Sound classes have been first introduced by \citet{Dolgopolsky1986}, who proposed 9 broad classes by which all possible consonants can be represented, searching for cognates across distantly related languages.
While this is a really crude reduction of phonetic detail, Dolgopolsky sound classes have been shown to work very well for comparative tasks \citep{Turchin2010}. In our approach, we use Dolgopolsky's original consonant classes and represent vowels by an additional symbol. 
 
The fact that our original data are provided in CLDF with standardized phonetic transcriptions is a great advantage when it comes to the conversion of phonetic strings to sound classes. Since sound class conversion routines are readily available for phonetic transcriptions that conform to the standard for IPA proposed in CLTS, converting the cognate sets in German, Dutch, and English to sound classes requires very few preprocessing operations.

\subsection{Trimming Word Forms}
\label{sec:trimming}
We experiment with two different representations of word forms, full forms and trimmed forms, where we automatically exclude endings. Full forms reflect the word forms as they are typically encountered in dictionaries (with nominative case for nouns in German and infinitive endings for verbs). Bare stems are typically used in historical language comparison in order to show how words were historically related before they were modified in the respective descendant languages by various morphological processes. In order to obtain bare stems from our cognate sets in German, English, and Dutch, we make use of the recently introduced technique for the \emph{trimming} of phonetic alignments \citep{blum_list2023}. With this technique, those sites (columns) in a multiple phonetic alignment that show an exceeding amount of gaps (sounds that do not have counterparts across all languages in the sample) are excluded from the alignment. Although not identical with manually prepared word stem representations, we find that applying this technique drastically reduces the amount of gaps in the multiple alignments, while at the same time successfully removing verb endings in our sample. Table \ref{tab:datarep} displays the representation of our data with full forms vs. bare stems sound class representation.

\subsection{Linear Discriminative Learning Model}

In a first step, we evaluated the LDL model on the cognate data of each language separately. The model is trained and tested on all 339 word forms. 
Phonological input cues are 4-gram, 3-gram and 2-gram chunks of sound classes, while multilingual word embeddings are representing the semantic vectors. In a second step, we train the model on a single language, i.e. creating a naive speaker of a language with zero exposure to other languages, and subsequently test the model on the cognate data from the target language. In doing so, we are replicating the setting of psycholinguistic studies but overcome the limitations of previous language exposure to exclusively focus on the predictiveness of historical sound classes as cues to mutual intelligibility.

\subsection{Implementation}
The experiments are implemented in the form of Python and Julia scripts. For sound class conversion, we used the LingPy Python package \citep{LingPy}. For the extraction of bare word stems through trimming, the LingRex package was used \citep{LingRex}. For the implementation of the LDL models, the Linear Discriminative Learner from the \emph{JudiLing} package (an implementation of DL in the \emph{Julia} programming language) was used \citep{JudiLing}. 
Data and code needed to replicate the experiments from this study are curated on GitHub (\url{https://github.com/digling/intelligibility}) and archived with Zenodo (\url{https://doi.org/10.5281/zenodo.10609356}). 
Detailed instructions on how to run the code are given in the repository.

\begin{table}[tb]
\centering
\resizebox{\columnwidth}{!}{%
\tabular{c}
\begin{tabular}{llll}
\hline
& \textbf{4-grams} & \textbf{3-grams} & \textbf{2-grams}\\
\hline
German  & 0.99  & 0.93  & 0.51  \\
Dutch   & 1.0     & 0.93  & 0.52  \\
English & 1.0     & 0.95  & 0.54  \\
\hline
\end{tabular}\\
(a) Training data (full words)
\\
\begin{tabular}{llll}
\hline
& \textbf{4-grams} & \textbf{3-grams} & \textbf{2-grams}\\
\hline
German  & 0.99  & 0.92 & 0.50  \\
Dutch   & 1.0     & 0.89  & 0.48 \\
English & 1.0     & 0.99  & 0.57  \\
\hline
\end{tabular} \\
(b) Training data (trimmed words) \\\endtabular}
\caption{Comprehension accuracies on full (a) and trimmed (b) training data. Top-1 candidate is taken into account to compute accuracies.}
\label{tab:training_trimmed}
\end{table}

\begin{table*}[tb]
\centering
\resizebox{\textwidth}{!}{%
\begin{tabular}{|l|lll|lll|}
\hline
\bfseries Language Pair & \multicolumn{3}{c}{\bfseries (a) Full Word Forms} & 
\multicolumn{3}{c|}{\bfseries (b) Trimmed Word Forms} \\\cline{2-7}
& \textbf{4-grams} & \textbf{3-grams} & \textbf{2-grams} &
\textbf{4-grams} & \textbf{3-grams} & \textbf{2-grams} \\
\hline
GER-DUT &0.57 (0.71) & 0.51 (0.68) & 0.28 (0.52) &
 0.81 (0.86) & 0.75 (0.86) & 0.39 (0.65)
\\
DUT-GER &  0.51 (0.67) & 0.48 (0.61) & 0.25 (0.48) & 
0.82 (0.83) & 0.75 (0.83) & 0.40 (0.67)
\\
GER-ENG  &  0.68 (0.75) & 0.62 (0.73) & 0.29 (0.53)
&
0.79 (0.85) & 0.75 (0.84) & 0.33 (0.59)
\\
ENG-GER & 0.48 (0.59)  &   0.46 (0.55) & 0.23 (0.45) &
0.60 (0.66)  &   0.59 (0.64) & 0.32 (0.53)
\\
DUT-ENG & 0.68 (0.75) & 0.6313 (0.72) & 0.31 (0.55) &
0.77 (0.84) & 0.71 (0.81) & 0.30 (0.60)
\\
ENG-DUT & 0.53 (0.64) & 0.50 (0.59) & 0.29 (0.49) &
0.60 (0.67) & 0.59 (0.64) & 0.35 (0.54)
\\
\hline
\end{tabular}}
\caption{Comprehension accuracies of multilingual models for comprehension for (a) full word forms and (b) trimmed word forms. Values without brackets indicate results when the top-1 candidate is considered to compute accuracies, values in brackets indidcate results when top-5 candidates are considered.}
\label{tab:cross_trimmed}
\end{table*}
\section{Evaluation}
\subsection{Evaluation on Individual Languages}

Table \ref{tab:training_trimmed}(a) displays the comprehension accuracies on the training data for full word forms. 
For the evaluation process  only the predicted meaning, the top-1 candidate, was considered.
The evaluation results suggest a good comprehension memory of the model when Dolgopolsky sound classes are provided as 4-gram or 3-gram chunks. If sound classes are fed into the model as 2-gram chunks we observe a substantial drop in accuracy, indicating a reduced discriminative power to predict a semantic vector \emph{$\hat{S}$} that is similar to the gold standard vector \emph{$S$} of the training language. Table \ref{tab:training_trimmed}(b) displays the evaluation results 
after trimming word forms.
Comprehension accuracy remains high for 4-gram and 3-gram chunks. Again, the accuracy drops substantially when 2-gram chunks are taken into account. 

\subsection{Evaluation Across Languages}

Table \ref{tab:cross_trimmed}(a) illustrates the result of the multilingual models for full word forms. The first column contains the training-test language pairs. Values without brackets indicate accuracies when the top-1 candidate was taken into account for evaluation, values in brackets indicate accuracies when the correct meaning among top-5 candidates was considered. 
Allowing the model to evaluate comprehension accuracy based on a set of top-5 candidates accounts for possible confusion of the target word form with similar word forms, giving the model room for multiple answers. The cross-linguistic comprehension results in Table \ref{tab:cross_trimmed}(a) unsurprisingly replicate the chunk size effect we have seen in our training models, with 4-gram chunks providing the best comprehension results. We observe the best comprehension results for the language pair Dutch-English with an accuracy of 68\% (75\% for an evaluation on top-5 candidates), followed by German-English and German-Dutch. The worst comprehension results are given for a training on English and a test on German cognates (see row 4 of Table \ref{tab:cross_trimmed}(a)). \citet{gooskens_swarte_2017} report a similar result for human participants, indicating that our LDL models show a human-like performance when assessing comprehension abilities across languages.
 
Table \ref{tab:cross_trimmed}(b) displays the comprehension accuracies after applying the trimming procedure. 
Trimming phonetic alignments results in a substantial rise of prediction accuracies with Dutch-German, German-Dutch and German-English providing the best comprehension results. Again, the language pair English-German shows the lowest comprehension accuracy, similar to human results \citep{gooskens_swarte_2017}. 

\section{Discussion and Conclusion}
In this study we presented a computer-assisted method to mutual intelligibility based on a model that captures the cognitive processes by which humans comprehend languages. We expanded the model with multilingual semantic vectors and multilingual sound classes. Our multilingual sound classes were predictive 
when a combination of at least 3 sound classes is given, indicating that knowing the order of sound classes allows the model to comprehend languages from the same language family. However, we observe an effect of the training language, with English being the least advantageous language in our setting and in the data of \citet{gooskens_swarte_2017} with human participants. We report a higher accuracy for German-English than German-Dutch, again in line with the human data of \citet{gooskens_swarte_2017}. If sound classes are trimmed, we find the opposite effect. The pair Dutch-English shows better comprehension accuracies than Dutch-German, again with the opposite picture for the trimmed version. From a language learning perspective, the change of direction, i.e. the better prediction for German-Dutch and Dutch-German after trimming would imply a certain morphological knowledge of speakers. Speakers of German or Dutch knowing verb endings and ignoring them purposefully have an advantage in comprehending English. Our proposed model does not only offer a new method for automatic testing of mutual intelligibility but shows clear similarities to data obtained from human participants, making it a useful cognitive tool for research on language comprehension.

\section*{Supplementary Material}
All data and code needed to replicate the experiments discussed in this study are curated on GitHub (\url{https://github.com/digling/intelligibility}) and archived with Zenodo (\url{https://doi.org/10.5281/zenodo.10609356}). The German cognate dataset is also curated on GitHub (\url{https://github.com/lexibank/germancognates}) and archived with Zenodo (\url{https://doi.org/10.5281/zenodo.10609476}).

\section*{Limitations}
While our model offers some fruitful results for further investigation of mutual intelligibility, the dataset we provided contains a limited amount of carefully selected historical cognates. It remains to be seen how the model would deal with a much larger set of random words. Moreover, we cannot account for other language families or other languages than German, Dutch and English. However, we see our modeling procedure as a starting point for assessing mutual intelligibility computationally. For that reason, limiting our data to historical cognates and three languages only is a necessary step. For a complete picture, more languages from the Germanic language family need to be tested and the results need to be compared with comprehension results for other language families.  

\section*{Ethics Statement}
This research does not involve human or animal data. No potential ethical conflict or conflict of interest was reported by the authors.

\section*{Acknowledgements}
This research was supported by the Max
Planck Society Research Grant \emph{CALC³} (JML, \url{https://digling.org}) and
the ERC Consolidator Grant \emph{ProduSemy} (JML, Grant No. 101044282, see \url{https://doi.org/10.3030/101044282}). Views and opinions expressed are however those of the authors only and
do not necessarily reflect those of the European
Union or the European Research Council Executive Agency (nor any other funding agencies involved). Neither the European Union nor the granting authority can be held responsible for them. We thank Maria Heitmeier and Harald Baayen for their valuable input regarding the computational models used in this study. 


\appendix




\begin{thebibliography}{30}
\expandafter\ifx\csname natexlab\endcsname\relax\def\natexlab#1{#1}\fi

\bibitem[{Anderson et~al.(2018)Anderson, Tresoldi, Chacon, Fehn, Walworth, Forkel, and List}]{Anderson2018}
Cormac Anderson, Tiago Tresoldi, Thiago~Costa Chacon, Anne-Maria Fehn, Mary Walworth, Robert Forkel, and Johann-Mattis List. 2018.
\newblock \href {https://doi.org/https://doi.org/10.2478/yplm-2018-0002} {{A} {C}ross-{L}inguistic {D}atabase of {P}honetic {T}ranscription {S}ystems}.
\newblock \emph{Yearbook of the Poznań Linguistic Meeting}, 4(1):21--53.

\bibitem[{Anderson et~al.(2023)Anderson, Tresoldi, Greenhill, Forkel, Gray, and List}]{Anderson2023a}
Cormac Anderson, Tiago Tresoldi, Simon~J. Greenhill, Robert Forkel, Russell~D. Gray, and Johann-Mattis List. 2023.
\newblock \href {https://doi.org/https://doi.org/10.1093/jole/lzad011} {Variation in phoneme inventories: quantifying the problem and improving comparability}.
\newblock \emph{Journal of Language Evolution}, 0(0):1--20.

\bibitem[{Baayen et~al.(2019)Baayen, Chuang, Shafaei-Bajestan, and Blevins}]{baayen_LDL2019}
R.~Harald Baayen, Yu-Ying Chuang, Elnaz Shafaei-Bajestan, and James~P. Blevins. 2019.
\newblock \href {https://doi.org/10.1155/2019/4895891} {The discriminative lexicon: A unified computational model for the lexicon and lexical processing in comprehension and production grounded not in (de)composition but in linear discriminative learning}.
\newblock \emph{Complexity}, 2019.

\bibitem[{Blum and List(2023)}]{blum_list2023}
Frederic Blum and Johann-Mattis List. 2023.
\newblock \href {https://aclanthology.org/2023.sigtyp-1.6.pdf} {{Trimming phonetic alignments improves the inference of sound correspondence patterns from multilingual wordlists}}.
\newblock In \emph{Proceedings of the 5th Workshop on Computational Typology and Multilingual NLP}, pages 52--64. Association for Computational Linguistics.

\bibitem[{Casad(1987)}]{Casad1987}
Eugene~H. Casad. 1987.
\newblock \emph{Dialect intelligibility testing}.
\newblock Summer Institute of Linguistics, Dallas.

\bibitem[{Chuang et~al.(2018)Chuang, Bell, Banke, and Baayen}]{chuangetal2021multilingual}
Yu-Ying Chuang, Melanie~J. Bell, Isabelle Banke, and R.~Harald Baayen. 2018.
\newblock \href {https://doi.org/10.1111/lang.12435} {Bilingual and multilingual mental lexicon: A modeling study with linear discriminative learning}.
\newblock \emph{Language Learning}, 71(S1):219--292.

\bibitem[{Chuang et~al.(2023)Chuang, Kang, Xuefeng, and Baayen}]{chuang2023vector}
Yu-Ying Chuang, Mihi Kang, Luo Xuefeng, and R.~Harald Baayen. 2023.
\newblock \href {https://doi.org/https://doi.org/10.4324/9781003159759} {Vector space morphology with linear discriminative learning}.
\newblock In Davide Crepaldi, editor, \emph{Linguistic Morphology in the Mind and Brain}, 1st edition, pages 17--248. Routledge, London.

\bibitem[{Dolgopolsky(1986)}]{Dolgopolsky1986}
Aharon~B. Dolgopolsky. 1986.
\newblock {A} probabilistic hypothesis concerning the oldest relationships among the language families of northern {E}urasia.
\newblock In Vitalij~V. Shevoroshkin, editor, \emph{{T}ypology, {R}elationship and {T}ime}, pages 27--50. Karoma Publisher, Ann Arbor.

\bibitem[{Forkel et~al.(2018)Forkel, List, Greenhill, Rzymski, Bank, Cysouw, Hammarström, Haspelmath, Kaiping, and Gray}]{Forkel2018a}
Robert Forkel, Johann-Mattis List, Simon~J. Greenhill, Christoph Rzymski, Sebastian Bank, Michael Cysouw, Harald Hammarström, Martin Haspelmath, Gereon~A. Kaiping, and Russell~D. Gray. 2018.
\newblock \href {https://doi.org/https://doi.org/10.1038/sdata.2018.205} {{C}ross-{L}inguistic {D}ata {F}ormats, advancing data sharing and re-use in comparative linguistics}.
\newblock \emph{Scientific Data}, 5(180205):1--10.

\bibitem[{Gooskens and Swarte(2017)}]{gooskens_swarte_2017}
Charlotte Gooskens and Femke Swarte. 2017.
\newblock \href {https://doi.org/10.1017/S0332586517000099} {{Linguistic and extra-linguistic predictors of mutual intelligibility between Germanic languages}}.
\newblock \emph{Nordic Journal of Linguistics}, 40(2):123–147.

\bibitem[{Gooskens et~al.(2015)Gooskens, van Bezooijen, and van Heuven}]{Gooskensetal2015}
Charlotte Gooskens, Renée van Bezooijen, and Vincent~J. van Heuven. 2015.
\newblock \href {https://doi.org/10.1515/ling-2015-0002} {{Mutual intelligibility of Dutch-German cognates by children: The devil is in the detail}}.
\newblock \emph{Linguistics}, 53(2):255--283.

\bibitem[{Gooskens et~al.(2018)Gooskens, van Heuven, Golubović, Schüppert, Swarte, and Voigt}]{gooskens2018mutual}
Charlotte Gooskens, Vincent~J. van Heuven, Jelena Golubović, Anja Schüppert, Femke Swarte, and Stefanie Voigt. 2018.
\newblock \href {https://doi.org/10.1080/14790718.2017.1350185} {{Mutual Intelligibility between Closely Related Languages in Europe}}.
\newblock \emph{International Journal of Multilingualism}, 15(2):169--193.

\bibitem[{Hammarström et~al.(2023)Hammarström, Haspelmath, Forkel, and Bank}]{Glottolog}
Harald Hammarström, Martin Haspelmath, Robert Forkel, and Sebastian Bank. 2023.
\newblock \href {http://arxiv.org/abs/https://glottolog.org} {\emph{{G}lottolog. {V}ersion 4.8}}.
\newblock Max Planck Institute for Evolutionary Anthropology, Leipzig.

\bibitem[{Heeringa et~al.(2014)Heeringa, Swarte, Schüppert, and Gooskens}]{heeringa_swarte_schüppert_gooskens_2014}
Wilbert Heeringa, Femke Swarte, Anja Schüppert, and Charlotte Gooskens. 2014.
\newblock \href {https://doi.org/10.1017/S1470542714000166} {{Modeling Intelligibility of Written Germanic Languages: Do We Need to Distinguish Between Orthographic Stem and Affix Variation?}}
\newblock \emph{Journal of Germanic Linguistics}, 26(4):361–394.

\bibitem[{Hongyan(2017)}]{Hongyan2017}
Wang Hongyan. 2017.
\newblock \emph{{English as a lingua franca: Mutual intelligibility of Chinese, Dutch and American speakers of English}}.
\newblock Ph.D. thesis, University of Utrecht.

\bibitem[{IPA(1999)}]{IPA1999}
International Phonetic~Organisation IPA. 1999.
\newblock \emph{{H}andbook of the {I}nternational {P}honetic {A}ssociation}.
\newblock Cambridge University Press, Cambridge.

\bibitem[{Kluge(2002)}]{Kluge2002}
Friedrich Kluge. 2002.
\newblock \emph{Etymologisches Wörterbuch der deutschen Sprache}, 24 edition.
\newblock de Gruyter, Berlin.

\bibitem[{List(2016)}]{List2016f}
Johann-Mattis List. 2016.
\newblock \href {https://doi.org/10.1093/jole/lzw006} {{B}eyond cognacy: {H}istorical relations between words and their implication for phylogenetic reconstruction}.
\newblock \emph{Journal of Language Evolution}, 1(2):119--136.

\bibitem[{List(2023)}]{EDICTOR}
Johann-Mattis List. 2023.
\newblock \href {http://arxiv.org/abs/https://digling.org/edictor} {\emph{EDICTOR. A web-based tool for creating, editing, and publishing etymological datasets [Software Tool, Version 2.1.0]}}.
\newblock MCL Chair at the University of Passau, Passau.

\bibitem[{List et~al.(2021)List, Anderson, Tresoldi, and Forkel}]{CLTS}
Johann-Mattis List, Cormac Anderson, Tiago Tresoldi, and Robert Forkel. 2021.
\newblock \href {https://doi.org/10.5281/zenodo.3515744} {\emph{{C}ross-{L}inguistic {T}ranscription {S}ystems. {V}ersion 2.1.0}}.
\newblock Max Planck Institute for the Science of Human History, Jena.

\bibitem[{List and Forkel(2023{\natexlab{a}})}]{LingPy}
Johann-Mattis List and Robert Forkel. 2023{\natexlab{a}}.
\newblock \href {https://pypi.org/project/lingpy} {\emph{{L}ing{P}y. {A} {P}ython library for quantitative tasks in historical linguistics {[Software Library, Version 2.6.13]}}}.
\newblock MCL Chair at the University of Passau, Passau.

\bibitem[{List and Forkel(2023{\natexlab{b}})}]{LingRex}
Johann-Mattis List and Robert Forkel. 2023{\natexlab{b}}.
\newblock \href {https://pypi.org/project/lingrex} {\emph{{L}ing{R}ex: {L}inguistic reconstruction with {L}ing{P}y}}.
\newblock Max Planck Institute for Evolutionary Anthropology, Leipzig.

\bibitem[{List et~al.(2022)List, Forkel, Greenhill, Rzymski, Englisch, and Gray}]{List2022e}
Johann-Mattis List, Robert Forkel, Simon~J. Greenhill, Christoph Rzymski, Johannes Englisch, and Russell~D. Gray. 2022.
\newblock \href {https://doi.org/10.1038/s41597-022-01432-0} {Lexibank, a public repository of standardized wordlists with computed phonological and lexical features}.
\newblock \emph{Scientific Data}, 9(316):1--31.

\bibitem[{Luo et~al.(2021)Luo, Chuang, and Baayen}]{JudiLing}
Xuefeng Luo, Yu-Ying Chuang, and R.~Harald Baayen. 2021.
\newblock \href {https://github.com/MegamindHenry/JudiLing.jl/tree/master} {{JudiLing: An implementation in Julia of linear discriminative learning algorithms for language modeling}}.

\bibitem[{Nieder et~al.(2023)Nieder, Chuang, van~de Vijver, and Baayen}]{niederetal_2023}
Jessica Nieder, Yu-Ying Chuang, Ruben van~de Vijver, and R.~Harald Baayen. 2023.
\newblock \href {https://doi.org/10.1353/lan.2023.a900087} {A discriminative lexicon approach to word comprehension, production, and processing: Maltese plurals}.
\newblock \emph{Language}, 99(2):242--274.

\bibitem[{Speer et~al.(2017)Speer, Chin, and Havasi}]{speer2017conceptnet}
Robyn Speer, Joshua Chin, and Catherine Havasi. 2017.
\newblock \href {http://aaai.org/ocs/index.php/AAAI/AAAI17/paper/view/14972} {{ConceptNet} 5.5: An open multilingual graph of general knowledge}.
\newblock In \emph{Proceedings of the AAAI Conference on Artificial Intelligence 2017}, pages 4444--4451.

\bibitem[{Squires et~al.(2020)Squires, Ohlfest, Santoro, and Roberts}]{Squires2020}
Lindsey~R. Squires, Sara~J. Ohlfest, Kristen~E. Santoro, and Jennifer~L. Roberts. 2020.
\newblock \href {https://doi.org/10.1044/2020_ajslp-19-00167} {Factors influencing cognate performance for young multilingual children’s vocabulary: A research synthesis}.
\newblock \emph{American Journal of Speech-Language Pathology}, 29(4):2170–2188.

\bibitem[{Tang and van Heuven(2007)}]{Tang2007}
Chaoju Tang and Vincent~J. van Heuven. 2007.
\newblock {M}utual intelligibility and similarity of {C}hinese dialects. {P}redicting judgments from objective measures.
\newblock In Bettelou Los and Marjo van Koppen, editors, \emph{{L}inguistics in the {N}etherlands 2007}, pages 223--234. John Benjamins Publishing Company.

\bibitem[{Tang and {van Heuven}(2009)}]{TANG2009709}
Chaoju Tang and Vincent~J. {van Heuven}. 2009.
\newblock \href {https://doi.org/https://doi.org/10.1016/j.lingua.2008.10.001} {{Mutual intelligibility of Chinese dialects experimentally tested}}.
\newblock \emph{Lingua}, 119(5):709--732.

\bibitem[{Turchin et~al.(2010)Turchin, Peiros, and Gell-Mann}]{Turchin2010}
Peter Turchin, Ilja Peiros, and Murray Gell-Mann. 2010.
\newblock {A}nalyzing genetic connections between languages by matching consonant classes.
\newblock \emph{Journal of Language Relationship}, 3:117--126.

\end{thebibliography}
\end{document}